\documentclass[journal, a4paper]{IEEEtran}

\usepackage[margin=0.5in]{geometry}

\usepackage{xcolor}
\usepackage{amssymb}
\usepackage{listings}
\usepackage{inconsolata}
\usepackage{caption}
\usepackage{tabularx}
\usepackage{float}
\usepackage{graphicx}   
\usepackage{url}        
\usepackage{amsmath}
\begin{document}

% Define document title, do NOT write author names
\title{An Adaptive Conversational Bot Framework}
\author{Isak Czeresnia Etinger  \textbf{[Microsoft][\'{E}cole Polytechnique]}
\\ \footnotesize isak.czeresnia-etinger@polytechnique.edu}
\maketitle

% Write abstract here
\begin{abstract}
How can we enable users to heavily specify criteria for database queries in a user-friendly way? This paper describes a general framework of a conversational bot that extracts meaningful information from user's sentences, that asks subsequent questions to complete missing information, and that adjusts its questions and information-extraction parameters for later conversations depending on users' behavior. Additionally, we provide a comparison of existing tools and give novel techniques to implement such framework. Finally, we exemplify the framework with a bot to query movies in a database, whose code is available for Microsoft employees. 
\end{abstract}

% Each section begins with a \section{title} command
\section{\textbf{Introduction}}

Consider the problem of recommending movies to users: it is a long-standing problem in data science that has implied a variety of techniques \cite{bib1}\cite{bib2}, ranging from Conditional Random Fields (for language understanding) to Collaborative Filtering techniques (for recommendation based on users' feedback on watched movies). It does not lack applications: movies digital stores abound, and their monetization greatly depends on how easily users can find movies they are interested in watching at particular moments.

Traditional solutions to movie search focused on non-iterative queries (e.g.: a one-time search on Bing) and at user-level preferences. The conversational bot framework that we propose introduces extra features: \\ 1) Its iterative nature improves user experience and facilitates the input of a greater number of criteria by the user. \\ 2) It enables users to query movies for criteria chosen anytime, which accounts for different preferences of a same user in different moments, and thus offers an advantage over the user-level preferences model. 

We implemented the proposed framework for a movie-searching bot that serves as example; but it can be generalized for any problem of conversational bots extracting criteria to query databases and adapting themselves to user behavior.   

\section{\textbf{General Bot Algorithm}}

The algorithm works by extracting \textit{entities} (section \ref{LanguageUnderstanding}) with a language understanding service, and using those entities to calculate indexes to be queried on a database (e.g.: of movies).
\vspace{-0.6 cm}

\begin{figure}[H]
	\centering
	\fbox{\includegraphics[width=1.0\columnwidth]{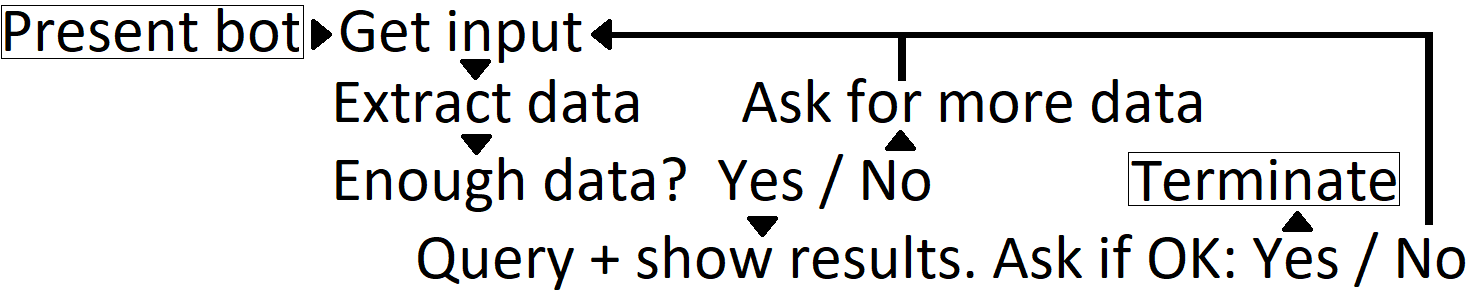}}
    \caption{\label{a_figure0}General bot structure}
\end{figure}

\section{\textbf{Data structure}} \label{DataStructure}
The movie database created is an Azure Document DB with 41482 documents (movies), each has the following properties (arrays \underline{underlined}):
Id (Microsoft's "Big Id"), title, release year, audience age, quality rating, \underline{genre ids}, \underline{directors}, \underline{actors}, \underline{directors \textit{Metaphone} keys}, \underline{actors \textit{Metaphone} keys}, \underline{countries}, \underline{continents}, main country, main continent.

\textit{Metaphone} keys are keys that represent how a word is pronounced. This is used in order to account for users mistyping the name of a director or actor, or for the voice understanding service transcripting a name differently than registered. E.g.: all "Steven Spielberg", "Steven Spilberg" and "Steven Spillberg" have the same Metaphone key "STFN SPLPRK". We used the Metaphone 2 algorithm \cite{bib3} to generate the keys. Alternatively, one could use the Metaphone 3 algorithm, which is more modern but is copyrighted and cannot be used freely.

\vspace{0.15 cm}

To register data extracted from the conversation, we created a \textit{Specification} class. Multiple instances of the class store each a type of specification extracted: genre, release year, audience age, country(ies)/continent(s) of origin, director(s)/actor(s) present, and extra keywords. The context of the conversation is solely composed of those 6 instances of the Specification class + the last question asked by the bot.

\lstset{
  basicstyle=\ttfamily,
}

Each instance of the Specification class is comprised of:
\begin{itemize}
\item a Dictionary$<$string, double$>$: the words extracted (of the Specification type) and their certainty scores.
\item a queue of bools: for each of the last K conversations, whether the specification was given.
\item a queue of ints: for each of the last K conversations, the order in which the specification was given (or refused).
\end{itemize}

Words and certainty scores from the Dictionary are used to perform queries on the database. The queues are used as training data for the language understanding. 

\section{\textbf{Language Understanding}} \label{LanguageUnderstanding}
Language understanding services operate by classifying an input phrase into \textit{intents} and by extracting \textit{entities} from it. Intents represent the intentions of the phrase. Entities represent terms or objects in the input that provide clarification or specification for a particular intent. Intents and entities returned by the service are accompanied by certainty scores $\in [0, 1]$.

We propose the use of only three intents: 
\begin{itemize}
\item Give specification;
\item Refuse to give the requested specification type.
\item None (neutral default intent)
\end{itemize}

And, for movies, the use of the entities (parentheses indicate the corresponding prebuilt entity in LUIS Entertainment class\cite{bib4}):
\begin{itemize}
\item Audience age (ContentRating)
\item Genre (Genre)
\item Keyword (Keyword)
\item Country/Continent of origin (Nationality + Language)
\item Director/Actor present (Person)
\item Release year
\end{itemize}

E.g., for any language understanding service we should have:
\vspace{-0.5 cm}

\begin{table}[H]
	\centering
    \tabcolsep=0.15cm
	\begin{tabular}{|llp{4.2cm}|}
		\hline
		\textbf{Input} & \textbf{Intent} & \textbf{Entities}  \\
		\hline
        "Hey"	& None	& 	$\emptyset$ \\
		"No, it doesn't matter"        & Refuse: 0.95        & $\emptyset$ \\
        "I want a comedy movie"        & Specify: 0.9        & Genre("comedy": 0.9) \\
        "Give me a comedy or action        & Specify: 0.9        &  Genre("comedy": 0.9, "action": 0.9),\\
        movie by Steven Spielberg" & & Person("Steven Spielberg": 0.8)\\
        \hline
	\end{tabular}
    \caption*{\small{Table 1: example of intent and entities extraction}}
\end{table}

The trivial method to use such a service is to consider returned intents and entities as valid iff their certainty score are higher than a fixed threshold (e.g.: 0.7), and to pass the retrieved certainty score unchanged to the Specification instances.

Note that this enables users to give specifications of types different than the one the bot just asked about. E.g.: If the bot asks "What genre of movies would you like?", and the user answers "It can be comedy or action, and by Steven Spielberg please", then the entity "Steven Spielberg" of type Person will also be stored. This has the advantage of giving more freedom to the user and improving user experience.
\vspace{0.15 cm}

We propose several techniques to increase performance over the traditional method by making the bot adapt to conversations. We divide those techniques into \textit{intra-conversation} (within a conversation) and \textit{inter-conversation} (between conversations) adaptations.

%\vspace{0.2 cm}

\subsection{\textbf{Intra-conversation Adaptations}}

\textbf{1. Asked entity-type bias:} consists of using a positive bias for certainty scores of entities of the type of the last question asked by the bot. For example, if the last question asked was "What genre of movie would you like?", then we could add a +0.2 bias to the certainty score of entities of the Genre type returned by the language understanding service, and then consider entities valid iff they had a total score $\geq$ 0.7. This way, entities of the Genre type would need a minimum certainty score of 0.5, while entities of other types would need the original 0.7 certainty score to be considered valid. The certainty score passed to the Specification instance should be the sum [service certainty score + bias]. This adaptation is simple to implement and accounts for users answering the bot's questions.

\vspace{0.2 cm}

\textbf{2. Word search for entity type:} consists of searching the input text for words of the entity type last asked by the bot. We propose to have a Dictionary mapping words of an entity type to indexes used on the database. For example, movie genres can be stored as integers in the database, and the Dictionary can contain a mapping \{"comedy":1, "funny":1, "to laugh":1, "action":2, "terror":3, "horror":3, ...\}. If the last question asked was "What genre of movie would you like?", and if no entities were found, then we can search the input text for words present in the Dictionary: "comedy", "funny", ...

The certainty score passed to the Specification instance should be decreasing according to the number of words in the input (e.g.: 1.0 for 1-worded inputs, 0.7 for 2-worded, ...). This method cannot be used for entity types lacking such indexing Dictionary (e.g.: for directors and actors names). For entities that have such indexing Dictionaries, this technique is easy to implement, and accounts for many cases in which the user answers a question with one or few words. Extremely short answers are problematic to language understanding services because grammatical techniques such as part-of-speech tagging (POS tagging) are not applicable (e.g.: if the input is "Spielberg", "Comedy", "France" or "Children", then grammatical analyses do not differentiate them -- all are single-noun phrases). This technique helps overcome that challenge.

\vspace{0.2 cm}

\textbf{3. Negative certainty scores:} consists of inverting the sign of an entity if it was found to be a negative specification. A complicate implementation is to apply a complete POS tagging and check connections of negation expressions (e.g.: not, except, but, anything other than, different from, ...) and the entity. A simple implementation is to check for negative expressions between entities. If an input consists of characters \{$A_0$, ..., $A_N$\}, and if entity $E_i$ is in positions \{$S_i$, ..., $T_i$\}, then the characters to be checked for entity $E_i$ are \{i $>$ 0 ? $T_{i-1} + 1$ : 0, ..., $S_i$ - 1\}. When querying on the database, negative certainty scores can be understood as negation, so this technique adds a powerful feature to the user specification ability.

\subsection{\textbf{Inter-conversation Adaptations}}
\textbf{4. Intra-type skip model:} consists of modeling the probability that a user will refuse to give an entity type (\textit{skip} a type) as depending only on the history of skips of that type in past conversations. For each type, for the last K (e.g.: 10) conversations, we store whether the type was skipped on the conversation, and we use those values to calculate the skip probability. One can use $\hat{p} = \frac{\# skips}{\# convReg}$ (where $\#convReg \leq$ K is the number of conversations registered), or use a weighted sum of skips where recent conversations have greater weight. One can also have an initial bias for when the number of conversations is less than K: $\hat{p} = \frac{\# skips \; + \; bias \times (K - \#convReg)}{K}$.

The estimated skipping probability for a type can be used to create a bias for a certainty score. E.g.: we can have a bias $b = - \alpha (\hat{p} - 0.5)$, or $b = - \alpha (\hat{p} - 0.5) ^ 3$, where $\alpha > 0$. Due to the negative sign, an entity type has greater certainty score iff it had a higher skip rate in past conversations. 

It is important not to have an excessively high value for $\alpha$, however. It is easy to see that $\hat{p}$ could converge to 0.0 or 1.0: the nearer $\hat{p}$ is to 1.0 (resp. 0.0), the more likely for the next conversation 
to be considered a skip (resp. not a skip).
\vspace{0.2 cm}

\textbf{5. Inter-type order-independent skip model:} consists of modeling the probability of skip as depending on the history of skips of all the types present. More formally, consider that the bot extracts entities of types \{$E_1$..$E_N$\}, and stores the skips of the past K conversations \{$C_1$..$C_K$\} ($C_1$ the newest). Then $\forall i \in \{1..N\},\; j \in \{1..K\}$, we have:

\vspace{0.15 cm}
\centerline{$\{true, false\} \ni s_{i,j}$ = (whether $E_i$ was skipped in $C_j$)}
\vspace{0.15 cm}

\noindent
Also consider that $I' = \{i' \,|\, s_{i',0}\mbox{  \footnotesize assumed}\}$ are the indexes of assumed skips of the current conversation. $s_{i',0}$ is \textit{\underline{assumed}} true if the user refused to specify $E_{i'}$ when asked; it is assumed false if $E_{i'}$ was already extracted in the conversation; and otherwise it is not assumed. The skip probability estimation then is, $\forall i \notin I'$,

\vspace{0.15 cm}
%\centerline{$\hat{p}_{i,0} = \hat{p}_{i,0}\Big(\{s_{i',0}\}_{i' \in I'} \cup \{s_{i,j}\}_{i \in \{1..N\}, j \in \{1..K\}}\Big)$}
\centerline{$\hat{p}_{i,0} = \hat{p}_{i,0}\Big(\{s_{i',0}\;|\; \footnotesize\mbox{$i' \in I'$}\normalsize\}  \cup \{s_{i,j} \;|\; \footnotesize\mbox{$i \in \{1..N\}, j \in \{1..K\}$}\}\Big)$}
\vspace{0.15 cm}

Note that this model does not differentiate the order in which entity types are extracted (or skipped). It just assign each of them to one of 3 states (assumed skipped, assumed not skipped, not assumed).

We propose to ignore $\{s_{i^*,j} \; | \; \mbox{\small$i^* \notin I' \wedge i^* \neq i \wedge j \in \{1..K\}$ \normalsize}\!\!\}$ (i.e.: the history of types that are neither assumed nor the type whose probability is being estimated)
\footnote{Note: another option would be using $\{s_{i^*,j} | i^* \notin I' \wedge j \in \{1..K\}$\} to train a model that does not use $\{s_{i^*,0} | i^* \notin I'\}$ as prediction parameters. We ignore those models.}
so we can reduce the probability estimation to the extensively studied problem of Supervised Learning. We have, $\forall i \notin I'$,

\vspace{0.15 cm}
%\centerline{$\hat{p}_{i,0} = \hat{p}_{i,0}\Big(\{s_{i,j}\}_{j \in \{1..K\}} \cup \{s_{i',j}\}_{i' \in I' \; \wedge \; j \in \{0..K\}}\Big)$}
\centerline{$\hat{p}_{i,0} = \hat{p}_{i,0}(\{s_{i,j} \;|\;\mbox{\footnotesize$j \in \{1..K\}$} \} \cup\{s_{i',j} \;|\;\mbox{\footnotesize$i' \in I'_0 \wedge \; j \in \{0..K\}$} \})$}
\vspace{0.15 cm}

In other words, we have the history of skips (including current conversation) for each of the assumed entity types, and also the history of skips (excluding current conversation) for the type whose probability we want to estimate. History (excluding current conversation) of $I'$ and $i$ serve as the training for the Supervised Learning regressor, then skip information of the current conversation $(\{s_{i',0} \;|\; i' \in I'\})$ serve as parameters for the regressor to estimate the probability.

\vspace{0.15 cm}

We now present a specific regressor for this problem which is easier to implement than most Supervised Learning algorithms.

\begin{table}[h]
\begin{center}
\resizebox{\linewidth}{!}{
\begin{tabular}{|l|}
\hline
function $\hat{p}_1(i, 0) = $\\
\qquad let function $D_0(j) = \#\{i' \in I' \;|\; s_{i',j} \neq s_{i',0}\}$\\
\qquad let set $Conv = \underset{j \in \{1..K\} \qquad\quad\;\;\; \textcolor{white}{.}}{argMin \; (D_0(j))}$ \normalsize\\
%\qquad let set $X = $ \Large $ argMin \; (D(i^*, i)) \atop i^* \in (\{1, ..., N\} \setminus \{i\}) $\normalsize \\ 
%\qquad let set $ X = \genfrac{}{}{0pt}{0}{argMin \; (D(i^*, i))}{\mbox{\footnotesize{$i^* \in (\{1, ..., N\} \setminus \{i\})$}}}$\\

%\qquad return \large $\underset{c \;\in\; Conv \textcolor{white}{.}}{average} \; \Big(\frac{\# \{i' \; assumed \;|\; s_{c,i'}\} }{\# \{i' \; assumed\}}\Big)$ \normalsize\\
%\qquad return \Large $\frac{\# \{(i',c) \;|\; c \;\in\; Conv \;\wedge\; s_{c,i'}\}}{\# \{(c,i') \;|\; c \;\in\; Conv\}}$ \normalsize \\
\qquad return \large $\frac{\# \{c \;\in\; Conv \;|\; s_{i, c}\}}{\# Conv}$ \normalsize \\
\hline
\end{tabular}
}
\end{center}
\vspace{-0.3 cm}
\caption*{\small{Algorithm 1: a simple estimator for skip probability}}
\end{table}

In other words, the algorithm defines a distance-from-current-conversation $D_0$ as the number of assumed entity types on which a conversation differs (skip-wise) from the current one. It selects the set of past conversations closest to the current one, and returns the fraction of those conversations in which the entity type $E_i$ was skipped.

\vspace{0.15 cm}

A possible disadvantage of that estimator is its low resiliency to outliers. Suppose that a single conversation (which did not skip $i$) had the same skips as the current one for all assumed entity types, and that many conversations (which all skipped $i$) differed  from the current one by exactly one assumed entity type (not necessarily all the same type). The first algorithm estimates $\hat{p} = 0.0$ because it only considers the closest conversation. Perhaps a more holistic algorithm taking into account neighboring conversations would be preferable. The following algorithm addresses that challenge ($\alpha > 0$ fixed):

\begin{table}[h]
\begin{center}
\resizebox{\linewidth}{!}{
\begin{tabular}{|l|}
\hline
function $\hat{p}_2(i, 0) = $\\
\qquad let function $D_0(j) = \#\{i' \in I' \;|\; s_{i',j} \neq s_{i',0}\}$\\
\qquad let function $Weight(j) = exp(-\;\alpha\; D_0(j)) $\\
%\footnotesize \textcolor{white}{.} \normalsize\\
\qquad return \large $\frac{\sum\nolimits_{j \;\in\; \{1..K\} \;|\; s_{i, j}} Weight(j)}
{\sum\nolimits_{j \;\in\; \{1..K\}} Weight(j)}$\\
%\footnotesize \textcolor{white}{.}\\
\hline
\end{tabular}
}
\end{center}
\vspace{-0.3 cm}
\caption*{\small{Algorithm 2: a more resilient estimator for skip probability}}
\end{table}

\vspace{-0.1 cm}

$Weight$ can be any arbitrary positive and decreasing function. The slower it decays, the smoother its behavior will be to outliers. If it sharply decays for all values, then $\hat{p}_2$ will be equivalent to $\hat{p}_1$. If it varies very little, then it will be equivalent to the intra-type model (because it will not differentiate conversations closer or more distant from the current one).

\vspace{0.15 cm}
\centerline{$\underset{\alpha \rightarrow \infty}{limit} \; \hat{p}_2(i, 0) = \hat{p}_1(i, 0)$}
\vspace{0.15 cm}
\centerline{$\underset{\alpha \rightarrow 0^+}{limit} \; \hat{p}_2(i, 0) = \hat{p}_{\mbox{\footnotesize intra-type model}}(i, 0)$}

\vspace{0.3 cm}

\textbf{6. Inter-type order-dependent skip model:} consists of modeling the probability of skip as depending on the order with which the entity types were assumed and whether they were skipped. We recall that an entity type skip $s_{i', 0}$ is assumed true if the user refused to specify the type when asked; it is assumed false if the type was already extracted in the conversation; and otherwise it is not assumed.

More formally, consider that the bot extracts entities of types \{$E_1$..$E_N$\}, and for each conversation $\in$ \{$C_1$..$C_K$\} ($C_1$ the newest) stores the order in which the types were assumed. Then $\forall i \in \{1..N\}, j \in \{1..K\}$, we have:

\vspace{0.15 cm}
\centerline{$o_{i,j} = \#\{i^* \in \{1..N\} \;|\; E_{i^*} \mbox{ was assumed before $E_i$ in $C_j$}   \} $}
\vspace{0.15 cm}

\noindent
Recall $I'$ is defined as the indexes of assumed skips of the current conversation. The skip probability estimation is, $\forall i \notin I'$,

\vspace{0.15 cm}
\centerline{$\hat{p}_{i,0} = \hat{p}_{i,0}( 
\{o_{i^*,j}, s_{i^*,j} \;|\; \mbox{\small$(i^*,j) \in ( (I' \!\! \times \!\! \{0\}) \cup (\{1..N\} \!\! \times \!\! \{1..K\})$\normalsize}) \})$}
\vspace{0.15 cm}

%\vspace{0.15 cm}
%\centerline{$\hat{p}_{i,0} = \hat{p}_{i,0}\Bigg( 
%\underset{i' \in I'}{\{o_{i',0}\}}, 
%\underset{i \in \{1..N\}, j \in \{1..K\}}{\{o_{i,j}\}}, 
%\underset{i' \in I'}{\{s_{i',0}\}}, 
%\underset{i \in \{1..N\}, j \in \{1..K\}}{\{s_{i,j}\}}\Bigg)$}
%\vspace{0.15 cm}

%\noindent
The \textbf{first technique (6.1)} we propose is to ignore $\{s_{i^*,j}, o_{i^*,j} \;|\; \mbox{\footnotesize $i^* \notin I' \;\wedge\; i^* \neq i$}\}$ as the order-independent model:

\vspace{0.15 cm}
\centerline{$\hat{p}_{i,0} = \hat{p}_{i,0}( 
\{s_{i^*,j}, o_{i^*,j} \;|\; \mbox{\small$(i^*,j) \in ( (\{i\} \!\! \times \!\! \{1..K\}) \cup (I' \!\! \times \!\! \{0..K\})$\normalsize}) \})$}
\vspace{0.15 cm}

%\vspace{0.15 cm}
%\centerline{$\hat{p}_{i,0} = \hat{p}_{i,0}\Bigg( 
%\underset{j \in \{1..K\}}{\{o_{i,j}\}}, 
%\underset{i' \in I', j \in \{0..K\}}{\{o_{i',j}\}}, 
%\underset{j \in \{1..K\}}{\{s_{i,j}\}}, 
%\underset{i' \in I', j \in \{0..K\}}{\{s_{i',j}\}}\Bigg)$}
%\vspace{0.15 cm}

%\noindent
Note that ignoring those values still keeps indirect information about the order of the ignored entity types. E.g.: in a conversation, types $(E_1, E_2, E_3)$ were assumed in order (2, 0, 1). If in the current conversation only $E_1$ was assumed and $E_2$ is being estimated, then we ignore the past order of $E_3$ (1). However, the past orders of $E_1$(2) and $E_2$(0) provide indirect information about the order of $E_3$ (they imply that $o_3=1$).

The problem is then analogous to the one presented in the order-independent model. It suffices to choose a different distance-from-current-conversation. We propose ($\beta > 0$  fixed):

\vspace{0.15 cm}
\centerline{$D_0(j) = \#\{i' \in I' \;|\; s_{i',j} \neq s_{i',0}\} + \beta \sum\nolimits_{i' \in I'} Abs(o_{i',j} - o_{i',0})$}
\vspace{0.15 cm}

\vspace{0.3 cm}
The \textbf{second technique (6.2)} we propose is for when the bot just asked about $E_i$ and we are estimating $s_{i,0}$. We ignore $\{s_{i^*,j} \;|\;\mbox{\footnotesize$ i^* \notin I' \;\wedge\; i^* \neq i$}\}$ and $\{o_{i^*,j} \;|\; i^* \notin I'\}$ like in 6.1, but we do not ignore $\{o_{i, j} \;|\;\mbox{\footnotesize$j \in \{1..K\}$}\}$, and we consider $o_{i,0} = \#I'$. All is analogous to 6.1. We have ($\gamma > 0$ fixed):

\vspace{0.15 cm}
\noindent
\mbox{$D_0(j) = \#\{i' \in I' \;|\; s_{i',j} \neq s_{i',0}\} + \beta \sum\nolimits_{i' \in I'} Abs(o_{i',j} - o_{i',0})$}
\hphantom {$D_0(j) = \#\{i' \in I' \;|\; s_{i',j} \neq s_{i',0}\}$} $ + \gamma Abs(o_{i,j} - \#I'$)
\vspace{0.15 cm}

\vspace{0.3 cm}

The \textbf{third technique (6.3)} we propose is to also use $\{s_{i^*,j}, o_{i^*,j} \;|\;\mbox{\footnotesize$ i^* \notin I' \wedge j \in \{1..K\}$}\}$ to train a model that does not use $\{s_{i^*,0}, o_{i^*,0} \;|\; i^* \notin I'\}$ as prediction parameters. When using the trained model to estimate $s_{i,0}$, we can use $\#I'$ to restrict the possible values of $\{o_{i^*,0} | i^* \notin I'\}$ -- e.g.: if 3 out of 5 entities were assumed in the current conversation, then we know that $(o_{4,0}, o_{5,0}) \in \{(3,4), (4,3), (3,3)\}$. In fact, the following algorithm calculates the set of possible orders. The notation used is similar to Python's List Comprehension: $(f(k))_{k \in \{1..A\}}$ represents a vector of size A $|$ element at position k is f(k).

\begin{table}[h]
\begin{center}
\resizebox{\linewidth}{!}{
\begin{tabular}{|l|}
\hline
function $OrdersSet()$ = \\
\quad let const int M = $N - \# I'$\\
\quad let set ($\subset \mathbb{N}^M$) Ret = $\emptyset$\\
\quad let set ($\subset \mathbb{N}^M$) Ords = $\{-1\}^M$\\
\quad while ($\exists \; O \in \mbox{Ords} \wedge k \in \{1..M\} \;|\; O_k = -1)$: \\
\quad \quad let set ($\subset \mathbb{N}^M$) NewOrds = $\emptyset$\\
\quad \quad foreach $int[M] \; \mbox{O} \in Ords$:\\
\quad \quad \quad let const int count = $\# \{k \in \{1..M\} \;|\; O_k \neq -1\}$\\
\quad \quad \quad foreach $E \in \{true, false\}^M \setminus \{false\}^M$:\\
\quad \quad \quad \quad if $\nexists \; k \in \{1..M\} \;|\; (E_k \wedge O_k \neq -1)$:\\
\quad \quad \quad \quad \quad let NewO = $(E_k \wedge O_K = -1 \;?\; count $ \\
\quad \quad \quad \quad \quad \hphantom{let NewO = } $: O_k)_{k \in \{1..M\}}$\\
\quad \quad \quad \quad \quad if NewO $\neq$ O $\wedge$ NewO $\notin$ Ret:\\
\quad \quad \quad \quad \quad \quad add NewO to NewOrds\\
\quad \quad \quad \quad \quad \quad if $(\mbox{NewO}_k \neq -1 \; \forall k \in \{1..M\})$:\\
\quad \quad \quad \quad \quad \quad \quad add $(\# I'+ \mbox{NewO}_k)_{k \in \{1..M\}}$ to Ret\\
\quad \quad Ords = NewOrds\\
\quad return Ret\\

\hline
\end{tabular}
}
\end{center}
\vspace{-0.3 cm}
\caption*{\small{Algorithm 3: set of possible orders}}
\end{table}

%Denote $o^k$ the order of the $k^{th}$ new entity type, $\forall k \in \{1..M\}$.
O and NewO are vectors $\in \mathbb{N}^M$ such that if their $k^{th}$ element is not $-1$, then it is the order of the $k^{th}$ new entity type; if it is $-1$, then it means the entity type was not assumed yet.\\ The algorithm works by simulating successive user inputs, starting from O$\in\!\{-1\}^M$ (no types assumed yet): each transformation of O to NewO is a simulated input. $E \in \{true, false\}^M \setminus \{false\}^M$ is the vector of assumptions ($E_k$ is true iff the $k^{th}$ new entity type is assumed). For each $E$, we transform O into NewO such that for each entity type assumed in $E$ but not assumed in $O$, we write its order with the variable $count$. %The time complexity $\in \mathcal{O}(M \times 2^M)$.

After calculating $OrderSet$, we estimate $\{o_{i^*,0} \in OrderSet\\ \;|\; i^* \notin I'\}$. The problem is then reduced to the vastly studied field of Machine Learning Classification (to classify $o_{i^*,0}$ in one element of $OrderSet$). We propose a specific classifier (where $\delta > 0$ is a fixed parameter):

\begin{table}[h]
\begin{center}
\resizebox{\linewidth}{!}{
\begin{tabular}{|l|}
\hline
function $EstimatedOrdersWeights(OrderSet) \!= $\\
\quad let function $\hat{o}_{i^*,k} = (i^* \in I' \;? \; o_{i^*,k} : OrderSet_{k,i})$\\
\quad let function $D(j, k) = \#\{i' \in I' \;|\; s_{i',j} \neq s_{i',k}\}$\\
%\quad let function $\hat{o}_{i^*,j} = ((j = 0 \;\wedge\; i^* \notin I') \;?\; OrderSet_i : o_{i^*,j})$\\
\quad \hphantom{let function }+ $\beta \sum\nolimits_{i^* \in \{1..N\}} Abs(o_{i^*,j} - \hat{o}_{i^*,k})$\\
\quad let function $Weight(k) = \sum\nolimits_{j \in \{1..K\}} exp (- \delta D(j,k))$\\
%\quad let set indexesMinLoss $\underset{k \in \{1..K\} \qquad\quad \textcolor{white}{.}}{argMin \; (Loss(k))}$\\
\quad let map m = $\{(OrderSet_k \rightarrow Weight(k)) \;|\; $\\
\quad \hphantom{let map m = } $k \! \in \! \{1..\#OrderSet\}\}$\\
\quad return m normalized (i.e.: $\sum \mbox{weights of m} = 1$)\\
\hline
\end{tabular}
}
\end{center}
\vspace{-0.3 cm}
\caption*{\small{Algorithm 4: classifier for $\{o_{i^*,0} \in OrderSet \;|\; i^* \notin I'\}$}}
\end{table}

\vspace{-0.15 cm}

Then one can calculate $\hat{p}(i, 0)$ for each order in the (order $\rightarrow$ weight) map returned by EstimatedOrdersWeights, and the final $\hat{p}(i, 0)$ will be the sum of those $\hat{p}(i, 0)$ multiplied by their respective weights. A disadvantage of that approach is that it requires calculating $\hat{p}(i, 0)$ $\#OrderSet$ times. We have $M ! \leq \#OrderSet \leq M^M$, so the computation required might be prohibitive. Another approach we propose is to select only the orders with the highest weight, and use only their $\hat{p}(i, 0)$ to calculate the average.

\vspace{0.3 cm}

\textbf{7. Entity value skip model:} consists of also using the value of extracted entities (e.g.: "comedy") instead of only types (e.g.: "genre"). The distance function can be binary, e.g.: $D(value_1, value_2) = (value_1 = value_2 \;?\; 1 : 0)$, or it can be gradual by considering semantics, e.g.: $D("action", "adventure") < D("action", "drama")$.

We present this technique for the sake of completeness. However, we recommend using it only after having implemented all else. This model suffers from sparsity if the binary distance is implemented, and it might be complicated to implement the gradual solution with semantics.

\vspace{0.3 cm}

\textbf{8. Users metadata aggregation:} consists of aggregating the input metadata (i.e.: skip/order/value information, so $\{s_{i, j}, o_{i, j} \;|\; i \in \{1..N\} \wedge j \in \{1..K\}\}$ or the entity values described in 7.) from different users, in order to apply one of the techniques above by treating conversations of multiple users as the past conversations $\{C_1..C_K\}$.

Entity values might yield worries about privacy. Skip and order information should not. We propose the following:
\begin{itemize}
\item DB \textit{Metadata} stores metadata from past conversations. Receives new data from bots when a conversation ends;
\item DB \textit{ProbEstimations} stores estimated skip probabilities (each for a given set of current conversation metadata) + their timestamps;
\item Bots get skip probability estimation based on current conversation metadata as follows:
\begin{itemize}
\item Bot searches on \textit{ProbEstimations} if the probability has been estimated with a recent enough timestamp.
\item If it was: use it.
\item If it was not: calculate it with \textit{Metadata}, and store the result in \textit{ProbEstimations}.
\end{itemize}
\end{itemize}

Estimations can be stored in ProbEstimations as a mapping: (current conversation metadata) $\rightarrow$ (timestamp, probability). Past conversations metadata can be stored either as a simple mapping: (complete conversation metadata) $\rightarrow$ (number of past conversations with this metadata), or as a Trie where each level is an user input (and values inside the nodes are the number of past conversations with the metadata).

\vspace{1.3 cm} % to break column. Else vspace 0.3cm

\textbf{Other models:} Additional features for extra models include: 
\begin{itemize}
\item For each input: the number of extracted entities of a given type, time taken for the user to start responding, time taken to send the response, whether "backspace" or "delete" was pressed, and speaking tone.
\item For each conversation: the time of the day, the day of the week, and how much time the user spent on the movie-selling app during this session (for the Movie example).
\item For each user: age, gender, location, time since opened account or installed app.
\end{itemize}

\section{\textbf{Language Understanding Services Comparison}}

We chose Microsoft's LUIS \cite{bib4} as the language understanding service, in part due to integration benefits, but also because of how it compared to other services.

In terms of costs and scalability, we have:
\begin{itemize}
\item Google's Dialogflow\cite{bib5}:
	\begin{itemize}
	\item Standard edition: free, limit of 180 text queries /min, 1k voice queries /day, 15k voice queries /month.
	\item Enterprise edition: 2\$ /1k text query, ~6\$ /1k voice query. Limit 600 text queries /min, 300 voice queries /min, 100k voice queries /day.
	\end{itemize}
\item IBM Watson\cite{bib5}: free first 10k queries /month. Then 2.5\$ / 1k queries.
\item Microsoft's LUIS: free 10k text queries /month, or unlimited 1.5\$ /1k text queries; 5.5\$ /1k voice queries.
\item Open source Wit.ai\cite{bib5}: free.
\end{itemize}

As we can see, LUIS is 0.5\$ cheaper /1k queries than Google's service (both for text and for voice queries), and 1\$ cheaper /1k text queries than IBM Watson. 
\vspace{0.2 cm}

In terms of intents and entities extraction accuracy, we have:
\begin{itemize}
\item Wit.ai: lacks prebuilt entities for movies classification. This means that this service would require a considerable training work. Additionally, sample bots provided and found online are too simple to test reliability against complicated sentences (e.g.: they were designed only to understand date and time).
\item Google's Dialogflow: failed on test case. In their \underline{sample bot for TV control}, test cases "find me a comedy movie"/"find me an action movie", it detected the prebuilt context "tv-search", but it was unable to assign comedy or action to the "movie-genre" parameter. In their \underline{sample bot for Video search}, test case "Find me a French comedy movie from Steven Spielberg", it detected the "search-video" context, but it failed to extract "comedy", "French" and "Steven Spielberg".
\item IBM Watson: failed on test case. Their \underline{sample bot for} \underline{Watson Assistant and Tone Analyzer} should detect "emotions, social tendencies, and writing styles". It was unable to detect negative sentences, as shown in figure \ref{a_figure}. If "I don't feel so good" is replaced by "I feel bad", then the Bot returns "I detected anger" instead of "I detected joy". They did not present prebuilt entities for movies.

\begin{figure}[H]
	\centering
	\fbox{\includegraphics[width=1.0\columnwidth]{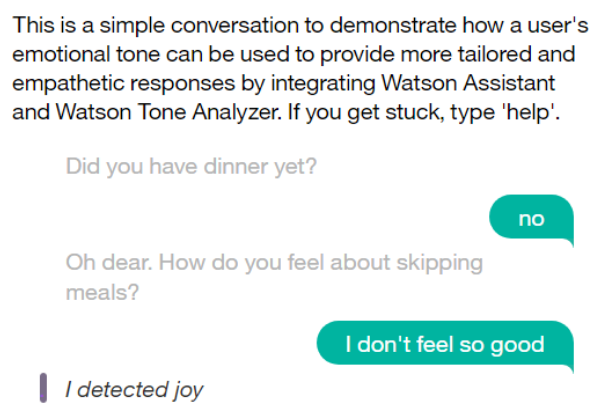}}
    \caption{\label{a_figure}IBM Watson failed test case. \small At: \protect\url{food-coach.ng.bluemix.net}}
\end{figure}

\end{itemize}

\section{\textbf{Asking Questions}}

After each user input, the bot asks the user for extra specifications, or, if enough information was already extracted: it queries on the database, shows top results to user and asks if the user is satisfied with results (at that point the user can either terminate the conversation or continue to give more specifications).

We propose that, in each question, the bot asks for the entity type which was not yet extracted and with the minimum estimated skip probability (i.e.: $argMin(\hat{p}_i)$) calculated with one or more techniques from the Language Understanding section. The bot considers the information extracted to be sufficient when all such estimated skip probabilities are above a certain threshold.

\section{\textbf{Querying the Database}}

To query the database (detailed in Data Structure section), the general use of the extracted entities is intuitive: for each entity type, add a "WHERE" clause to the SQL query specifying that a property of the movie has to be one of the entity values extracted of the type. Ignore entity types without values.

For the specific case of the movie finder, we have for each movie: origin countries (array), continents (array), main country and main continent. We consider the arrays only on matching (NOT)values, and consider main country/continent when matching usual values. We do this because many movies were originated by multiple countries but are only typical to the country listed as their main country.

Additionally, Metaphone keys are used for directors' and actors' names (as explained in Data Structure section). In the conversation sample of figure 2, the bot calculates the Metaphone key of the mistyped (or miss-transcripted from voice input) "Nataly Portman" into "NTL PRTMN". It is able to find the correct movie regardless of the "Natalie" typo.

There can be a second database: map PersonKey $\rightarrow$ List$<$movies$>$, such that finding a List$<$movies$>$ for a given PersonKey is an $\mathcal{O}(1)$ operation. The List$<$movies$>$ should contain the full information about the movies, not their Ids. If at least one actor/director is extracted, then the bot should query their keys in this database, and then apply additional WHERE clauses on the resulting List$<$movies$>$. This would be extremely fast and scalable to querying. A problem is the extra space required. If on average each movie has 20 actors/directors, then total stored data would be multiplied by \texttildelow 20. For the sample bot (\texttildelow41k movies), it should be worthwhile.

%\section{\textbf{Dataset}}
\begin{figure}[H]
	\centering
	\fbox{\includegraphics[width=1.0\columnwidth]{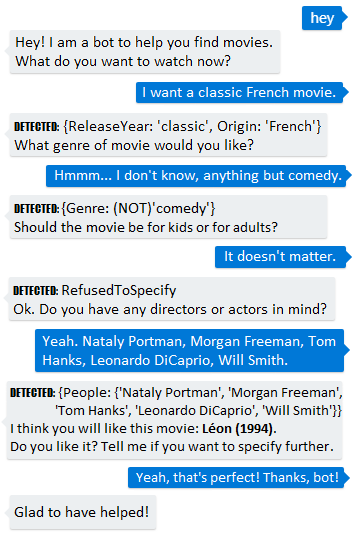}}
    \caption{\label{a_figure2}Movie finder sample conversation.}
\end{figure}

\section{\textbf{Conclusion}}
In this paper we presented a general framework to enable users to heavily specify criteria for database queries in a user-friendly way. It produces a conversational bot that extracts meaningful information from user's sentences, that asks subsequent questions to complete missing information, and that adjusts its questions and information-extraction parameters for later conversations depending on users' behavior. Additionally, we provided a comparison of existing tools and gave novel techniques to implement such framework.

We recommend developers to use LUIS as the language understanding service. For the language understanding techniques, we recommend developers to start by applying techniques 1-3 of the Language Understanding section, and then to gradually advance to technique 8 according to comfortability implementing the methods and tuning the parameters. Techniques 1-6 were implemented and are available to Microsoft employees.

% The bibliography:

% Your document ends here!
\end{document}